\documentclass[11pt,a4paper]{article}
\usepackage[hyperref]{emnlp2018}
\usepackage{braket} 
\usepackage{placeins}
\usepackage{mathtools}
\usepackage{times}
\usepackage{latexsym}
\usepackage{color}
\usepackage{amsmath}
\usepackage{amssymb}
\usepackage{url}
\usepackage{float}
\usepackage{bigstrut}
\usepackage[linewidth=1pt]{mdframed}
\usepackage{framed}
\usepackage{graphicx}
\usepackage{adjustbox}
\usepackage{array}
\usepackage{enumitem}
\usepackage{listings}
\usepackage{epstopdf}
\usepackage{framed}
\usepackage{cmll}
\usepackage{inputenc}
\usepackage{makecell}
\usepackage{graphicx}
\usepackage{subcaption}
\usepackage{booktabs}
\usepackage{xstring}
\usepackage{calc}
\usepackage{mdframed}
\usepackage{multicol}
\usepackage{boxedminipage}
\usepackage{tcolorbox}
\usepackage{etoolbox}
\usepackage{anyfontsize}
\usepackage{xcolor,colortbl}
\usepackage[ruled]{algorithm2e}
\usepackage{placeins}

\usepackage{multirow,hhline,graphicx,array}
\newcolumntype{M}[1]{>{\centering\arraybackslash}m{#1}}

\definecolor{commentgreen}{RGB}{2,112,10}
\definecolor{eminence}{RGB}{108,48,130}
\definecolor{weborange}{RGB}{255,165,0}
\definecolor{frenchplum}{RGB}{129,20,83}

\usepackage{listings}
\aclfinalcopy 

\newcommand{\namecite}[1]{\citeauthor{#1}~\shortcite{#1}}

\newcolumntype{L}[1]{>{\raggedright\let\newline\\\arraybackslash\hspace{0pt}}m{#1}}
\newcolumntype{C}[1]{>{\centering\let\newline\\\arraybackslash\hspace{0pt}}m{#1}}

\newcommand{\ignore}[1]{}
\definecolor{purple}{rgb}{0.5,0,1}
\definecolor{dcyan}{rgb}{0.2,0.6,0.5}
\definecolor{darkgreen}{rgb}{0,0.25,0}
\definecolor{light-gray}{gray}{0.95} 

\definecolor{DarkRed}{RGB}{130,25,0}

\title{
\vspace*{-0.5in}
{{\small \hfill EMNLP'18}\\
\vspace*{.25in}}
Zero-Shot Open Entity Typing as Type-Compatible Grounding}

\newcommand{\otyper}{\textsc{OTyper}}
\newcommand{\ours}{\textsc{Zoe}}  
\newcommand{\wikilinks}{\textsc{WikiLinks}}
\newcommand{\elmo}{\textsc{ELMo}}
\newcommand{\esa}{\textsc{ESA}}
\newcommand{\freebase}{\textsc{FreeBase}}
\newcommand{\wordnet}{\textsc{WordNet}}
\newcommand{\wikifier}{\textsc{Wikifier}}
\newcommand{\afet}{\textsc{AFET}}
\newcommand{\attentive}{\textsc{Attentive}}
\newcommand{\nfetc}{\textsc{NFETC-hier(r)}}
\newcommand{\protole}{\textsc{ProtoLE}}
\newcommand{\abishek}{\textsc{AAA}}

\newcommand{\cogcompnlp}{{\sc CogCompNLP}}

\newcommand{\corenlp}{{\sc CoreNLP}}
\newcommand{\spacy}{{\sc SpaCy}}

\newcommand{\nltk}{{\sc NLTK}}

\newcommand{\figer}{{FIGER}}
\newcommand{\bbn}{{BBN}}
\newcommand{\ontonotes}{OntoNotes}
\newcommand{\ontonotesFine}{OntoNotes$_{\text{fine}}$}
\newcommand{\conll}{CoNLL}
\newcommand{\muc}{MUC}
\newcommand{\bionlp}{BB3}

\newcommand{\accuracy}{Acc.}
\newcommand{\fOneMa}{$F1_{\text{ma}}$}
\newcommand{\fOneMi}{$F1_{\text{mi}}$}
\newcommand{\fOneMat}{$F1^{\text{type}}_{\text{ma}}$}
\newcommand{\fOneMit}{$F1^{\text{type}}_{\text{mi}}$}

\usepackage{booktabs,arydshln}

\makeatletter
\def\adl@drawiv#1#2#3{%
        \hskip.5\tabcolsep
        \xleaders#3{#2.5\@tempdimb #1{1}#2.5\@tempdimb}%
                #2\z@ plus1fil minus1fil\relax
        \hskip.5\tabcolsep}
\newcommand{\cdashlinelr}[1]{%
  \noalign{\vskip\aboverulesep
           \global\let\@dashdrawstore\adl@draw
           \global\let\adl@draw\adl@drawiv}
  \cdashline{#1}
  \noalign{\global\let\adl@draw\@dashdrawstore
           \vskip\belowrulesep}}
\makeatother

\author{Ben Zhou$^{1}$, Daniel Khashabi$^{2}$,  Chen-Tse Tsai$^{3}$, Dan Roth$^{2}$  \\
$^{1}$University of Illinois, Urbana-Champaign, \hspace{.2cm} $^{2}$University of Pennsylvania, \hspace{.2cm} $^{3}$Bloomberg LP\\
  {\tt \small xzhou45@illinois.edu,  \{danielkh,danroth\}@cis.upenn.edu, ctsai54@bloomberg.net} 
}

\date{}

\newcolumntype{R}[2]{
    >{\adjustbox{angle=#1,lap=\width-(#2)}\bgroup}
    l
    <{\egroup}
}

\definecolor{graaay}{rgb}{0.9, 0.9, 0.9}

\begin{document}
\maketitle
\begin{abstract}
The problem of entity-typing has been studied predominantly in supervised learning fashion, mostly with task-specific annotations (for coarse types) and sometimes with distant supervision (for fine types).  
While such approaches have strong performance within datasets, they often lack the flexibility to transfer across text genres and to generalize to new type taxonomies. 
In this work we propose a \emph{zero-shot} entity typing approach that requires no annotated data and can flexibly identify newly defined types. 


Given a {\em type taxonomy} defined as Boolean functions of \freebase\ ``types",
we ground a given mention to a set of {\em type-compatible} Wikipedia entries and then infer the target mention's types using an inference algorithm that makes use of the types of these entries.
%
We evaluate our system on a broad range of datasets, including standard fine-grained and coarse-grained entity typing datasets, and also a dataset in the biological domain. 
Our system is shown to be competitive with state-of-the-art supervised NER systems and outperforms them on out-of-domain datasets. We also show that our system significantly outperforms other zero-shot fine typing systems.

\ignore{
For the problem of entity-typing, the community has predominantly focused on the \emph{supervised} paradigm. 
While such approaches have strong performance within datasets, they often lack both the flexibility to transfer across writing genres 
and to work with new type taxonomies. In this work we propose a \emph{zero-shot} entity typing approach. 
Our system does not need to be trained on supervisions of target entity types. 
The key idea is to ground a given mention to a set of type-compatible Wikipedia entries, and then infer the target types based on 
the resulting Wikipedia entries and the given target type definitions.
We evaluate our system on a wide range of datasets, including both fine-grained and coarse-grained entity typing, and biological entity typing. 
Empirically we investigate our system's behavior compared to other zero-shot and fully supervised systems.
We show that our system outperforms a recently published zero-shot entity-typer by +7\% $F1$ score. In addition, our system achieves competitive results compared to the state-of-the-art supervised systems. }

\end{abstract}

\setlength{\abovedisplayskip}{4.4pt}
\setlength{\belowdisplayskip}{4.4pt}

\section{Introduction}
\label{sec:intro}
Entity type classification is the task of connecting an entity mention to a given set of semantic types. The commonly used type sets 
range in size and level of granularity, from a small number of coarse-grained types \cite{tjong2003introduction} to over a hundred fine-grained types \cite{ling2012fine}.
It is understood that 
semantic typing is a key component in many natural language understanding tasks, including Question Answering~\cite{toral2005improving,LiRo05a} and Textual Entailment~\cite{dagan2010recognizing,DRSZ13}. Consequently, the ability to type mentions semantically across domains and text genres, and to use a flexible type hierarchy, is essential for solving many important challenges.

Nevertheless, most commonly used approaches and systems for semantic typing
(e.g., \corenlp~\cite{manning2014stanford},  \cogcompnlp~\cite{cogcompnlp2018lrec}, \nltk~\cite{Loper2002nltk}, \spacy) are trained in a 
supervised fashion and rely on high quality, task-specific annotation.
Scaling such systems to other domains and to a larger 
set of entity types faces fundamental restrictions. %

Coarse typing systems, which are mostly fully supervised, are known to fit a single dataset very well. However, their performance drops significantly on different text genres and even new data sets. Moreover, adding a new coarse type requires manual annotation and retraining.  
For fine-typing systems, people have adopted a distant-supervision approach. 
Nevertheless, the number of types used is small: the distantly-supervised \figer\ dataset covers only $113$ types, a small fraction of most-conservative estimates of the number of types in the English language (the \freebase~\cite{freebase} and \wordnet~\cite{miller1995wordnet} hierarchies consist of more than $1k$ and $1.5k$ unique types, respectively).  
More importantly, adapting these systems, once trained, to new type taxonomies cannot be done flexibly.

As was argued in \newcite{Roth17}, there is a need to develop new training paradigms that support scalable 
semantic processing; specifically, there is a need to scale semantic typing to flexible type taxonomies and to multiple domains. 

\begin{figure*}
    \centering
    \includegraphics[trim=0cm 0cm 0cm 0.5cm, clip=false, scale=0.33]{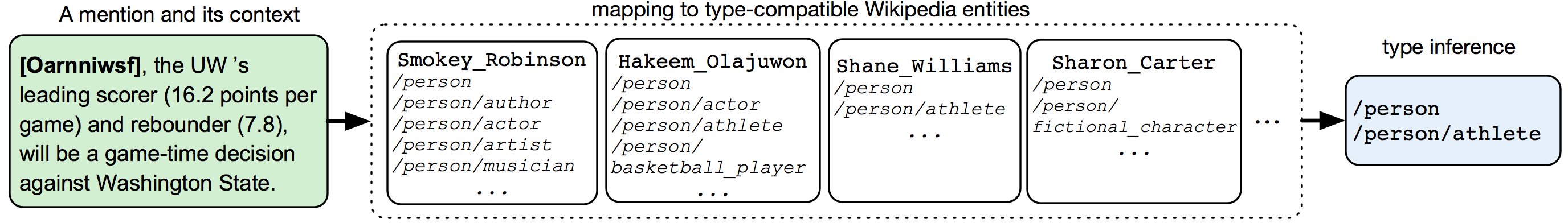}
    \vspace{-0.2cm}
    \caption{
    \ours\ maps a given mention to its type-compatible entities in Wikipedia and infers a collection of types using this set of entities. While the mention ``Oarnniwsf," a football player in the U. of Washington, does not exist in Wikipedia, we ground it to other entities with approximately the same types (\S\ref{sec:model}).
    }
    \vspace{-0.1cm}
    \label{fig:intro_example}
\end{figure*}

In this work, we introduce \ours, a \emph{zero-shot} entity typing system, with \emph{open} type definitions.
Given a mention in a sentence and a taxonomy of entity types with their definitions, \ours\ identifies a set of 
types that are appropriate for the mention in this context.
\ours\ does not require any training, and it makes use of existing data resources (e.g., Wikipedia) and tools developed without any task-specific annotation.
The key idea is to ground each mention to a set of \emph{type-compatible} Wikipedia entities. 
The benefit of using a set of Wikipedia titles as an intermediate representation for a mention is that there is much human-curated information in Wikipedia -- categories associated with each page, \freebase\ types, and DBpedia types. These were put there independently of the task at hand and can be harnessed for many tasks: in particular, for determining the semantic types of a given mention in its context.
In this grounding step, the guiding principle is that type-compatible entities often appear in similar contexts. 
We rely on contextual signals and, when available, surface forms, to rank Wikipedia titles and choose those that are more compatible with a given mention.

Importantly, our algorithm does not require a given mention to be in Wikipedia; in fact, in many cases (such as nominal mentions) the mentions are not available in Wikipedia. 
We hypothesize that any entity possible in English corresponds to some type-compatible entities in Wikipedia. We can then rely mostly on the context to reveal a {\em set} of {\em compatible titles}, those that are likely to share semantic types with the target mention.
The fact that our system is not required to ground to the exact concept is a key difference between our grounding and ``standard" Wikification approaches~\cite{mihalcea2007wikify,RRDA11}. As a consequence, while entity linking approaches rely heavily on priors associated with the surface forms and do not consider those that do not link to Wikipedia titles, our system mostly relies on context, regardless of whether the grounding actually exists or not. 

Figure~\ref{fig:intro_example} shows a high-level visualization of our system. 
Given a mention, our system grounds it into type-compatible entries in Wikipedia. The target mention ``Oarnniwsf," 
is not in Wikipedia, yet it is grounded to entities with approximately correct types. In addition, while some of the grounded Wikipedia entries are inaccurate in terms of entity types,  
the resulting aggregated decision is correct.

\ours\ is an \emph{open} type system, since it is not restricted to a closed set of types.
In our experiments, we build on \freebase\ types as {\em primitive} types and use them to define types across seven different datasets.
Note, however, that our approach is not fundamentally restricted to \freebase~ types; in particular, we allow types to be defined as Boolean formulae over these primitives (considering a type to be a set of entities). Furthermore, we support other primitives, e.g., DBPedia or Wikipedia entries. 
Consequently, our system can be used across type taxonomies; there is no need to restrict to previously observed types or retrain with annotations of new types. If one wants to use types that are outside our current vocabulary, one only needs to define the target type taxonomy in terms of the primitives used in this work.

\ignore{
In our experiments, we use \freebase\ types to define the types across seven different datasets.
Note, however, that our our system is fundamentally not restricted to the types in \freebase; one can combine the existing types in order to formalize a new type that does not exist in \freebase\ (even further, use DBPedia or Wikipedia categories, instead of \freebase\ entries). 
This removes the restriction to a closed set of observed labels in supervised systems. 
As a consequence, when moving across type taxonomies, there is no need to retrain our system on a dataset with annotations of new types. Rather one only has to give definition of the type taxonomy.
Intuitively, mapping type taxonomy to the information in Wikipedia is much cheaper than annotating hundreds of thousands of training examples. }

In summary, our contributions are as follows:
\begin{itemize}[nolistsep,topsep=1pt]
    \item We propose a zero-shot open entity typing framework\footnote{\url{https://github.com/CogComp/zoe}} that does not require training on entity-typing-specific supervised data.
    \item The proposed system outperforms existing zero-shot entity typing systems.  
    \item Our system is competitive  with fully-supervised systems in their respective domains across a broad range of coarse- and fine-grained typing datasets, and it outperforms these systems in out-of-domain settings. 
\end{itemize}


\section{Related Work}
\label{sec:related}
    


Named Entity Recognition~(NER), for which the goal is to discover mention-boundaries in addition to typing, often using a small set of mutually exclusive types, has a considerable amount of work~\cite{grishman1996message,mikheev1999named,tjong2003introduction,florian2003named,RatinovRo09}. 

There have been many proposals to scale the systems to support a bigger type space~\cite{Fleischman2002FineGC,sekine2002extended}. 
This direction was followed by the introduction of datasets with large label-sets, either manually annotated like \bbn~\cite{weischedel2005bbn} or distantly supervised like \figer~\cite{ling2012fine}. 
With larger datasets available, supervised-learning systems were proposed to learn from the data \cite{yosef2012hyena,abhishek2017fineGrained,shimaoka2017neural,xu2018neural,Choi2018UltraFineET}. 
Such systems have achieved remarkable success, mostly when restricted to their observed domain and labels. 

There is a handful of works aiming to pave the road towards zero-shot typing by addressing ways to extract cheap signals, often to help the supervised algorithms: e.g., by generating gazetteers~\cite{nadeau2006unsupervised}, or using the anchor texts in Wikipedia~\cite{nothman2008transforming,nothman2009analysing}. 
\namecite{Ren2016AFETAF} project labels in high-dimensional space and use label correlations to suppress noise and better model their relations. 
In our work, we choose not to use the supervised-learning paradigm and instead merely rely on a general entity linking corpus and the signals in Wikipedia.  
Prior work has already shown the importance of Wikipedia information for NER. 
\newcite{TsaiMaRo16} use a cross-lingual \wikifier\ to facilitate cross-lingual NER. However, they do not explicitly address the case where the target entity does not exist in Wikipedia.

The zero-shot paradigm for entity typing has only recently been studied. 
\namecite{yogatama2015embedding} proposed an embedding representation for user-defined features and labels, which facilitates information sharing among labels and reduces the dependence on the labels observed in the training set. The work of \namecite{yuan2018otyper} can also be seen in the same spirit, i.e., systems that rely on a form of representation of the labels. 
In a broader sense, such works--including ours--are part of a more general line of work on \emph{zero-shot learning}~\cite{Chang2008ImportanceOS,Palatucci2009ZeroshotLW,Norouzi2013ZeroShotLB,RomeraParedes2015AnES,Song2014OnDH}. 
Our work can be thought of as the continuation of the same research direction. 

\begin{table}[]
    \footnotesize
    \centering
\resizebox{\linewidth}{!}{
    \begin{tabular}{C{3.35cm}C{2.9cm}C{1.5cm}}
    \toprule
        Approach & Zero-shot? & Use labeled data? \\
    \midrule
        \shortstack{ \attentive \\ \cite{shimaoka2017neural} }  & No  &  Yes\\ 
    \cmidrule[0.1pt]{1-3}  
        \shortstack{ \abishek \\ \cite{abhishek2017fineGrained} } & No & Yes \\ 
    \cmidrule[0.1pt]{1-3}  
        \shortstack{ \nfetc \\ \cite{xu2018neural} } & No & Yes \\ 
    \cmidrule[0.1pt]{1-3}  
        \shortstack{ \afet \\ \cite{Ren2016AFETAF}} & No & \shortstack{Yes \\ (partial)} \\ 
    \cmidrule[0.1pt]{1-3}  
        \shortstack{ \protole \\  \cite{ma2016label} } & \shortstack{\textbf{Yes} \\{\footnotesize Prototype Embedding}} & \shortstack{ Yes \\ (partial) }\\ 
    \cmidrule[0.1pt]{1-3}  
        \shortstack{ \otyper \\ \cite{yuan2018otyper} } & \shortstack{\textbf{Yes} \\{\small Word Embedding}} & \shortstack{Yes\\ (partial)} \\ 
    \cmidrule[0.1pt]{1-3}  
        \shortstack{\cite{huang2016building} } & \shortstack{\textbf{Yes} \\{\small Concept-embedding} \\{\small Clustering}} & \textbf{No} \\ 
    \cmidrule[0.1pt]{1-3}  
        \ours~(ours) & \shortstack{\textbf{Yes} \\{\small Type-Compatible} \\{\small Concepts}} & \textbf{No} \\ 
    \bottomrule
    \end{tabular}
}
\vspace{-0.25cm}
    \caption{Comparison of recent work on entity typing. Our system does not require any labeled data for entity typing; therefore it works on new datasets without retraining.}
    \vspace{-0.1cm}
    \label{tab:comparison}
\end{table}

A critical step in the design of zero-shot systems is the characterization of the output space. For supervised systems, the output representations are trivial, as they are just indices. For zero-shot systems, the output space is often represented in a high-dimensional space that encodes the semantics of the labels. In \otyper~\cite{yuan2018otyper}, each type embedding is computed by  averaging the word embeddings of the words comprising the type. The same idea is also used in \protole~\cite{ma2016label}, except that averaging is done only for a few prototypical instances of each type. 
In our work, we choose to define types using information in Wikipedia. This flexibility allows our system to perform well across several datasets without retraining.
On a conceptual level, the work of \namecite{lin2012no} 
and \namecite{huang2016building} are close to our approach. The governing idea in these works  
is to cluster mentions, followed by propagating type information from representative mentions. 

Table~\ref{tab:comparison} compares our proposed system with several recently proposed models.

\newtheorem{definition}{Definition}
\newtheorem{hypothesis}{Hypothesis}

\section{Zero-Shot Open Entity Typing}
\label{sec:model}
Types are conceptual containers that bind entities together to form a coherent group. 
Among the entities of the same type, \emph{type-compatibility} creates a network of loosely connected entities: 

\begin{definition}[Weak Type Compatibility]
Two entities are type-compatible if they share at least one type with respect to a type taxonomy and the contexts in which they appear. 
\end{definition}

In our approach, given a mention in a sentence, we aim to discover type-compatible entities in Wikipedia and then infer the mention's types using all the type-compatible entities together. The advantage of using Wikipedia entries is that the rich information associated with them allows us to infer the types more easily. Note that this problem is different from the standard entity linking or Wikification problem in which the goal is to find the corresponding entity in Wikipedia. Wikipedia does not contain all entities in the world, but an entity is likely to have at least one type-compatible entity in Wikipedia.

In order to find the type-compatible entities, we use the context of mentions as a proxy. Defining it formally: 

\begin{definition}[Context Consistency]
A mention $m$ (in a context sentence $s$) is \emph{context-consistent} with another well-defined mention $m'$, if $m$ can be replaced by $m'$ in the context $s$, 
and the new sentence still makes logical sense. 
\end{definition}

\begin{hypothesis}
Context consistency is a strong proxy for type compatibility. 
\end{hypothesis}

Based on this hypothesis, given a mention $m$ in a sentence $s$, we find other context-compatible mentions in a Wikified corpus. 
Since the mentions in the Wikified corpus are linked to the corresponding Wikipedia entries, we can infer $m$'s types by aggregating information associated with these Wikipedia entries.

\definecolor{lightblue2}{RGB}{215, 231, 247}
\definecolor{lightpurple}{RGB}{255, 210, 251}
\definecolor{lightred2}{RGB}{255, 220, 216}
\definecolor{lightyellow}{RGB}{253, 255, 200}
\definecolor{lightgreen}{RGB}{213, 255, 211}

\begin{figure}
    \centering
    \includegraphics[trim=.0cm 0.5cm 0cm 0.4cm, clip=false, scale=0.30]{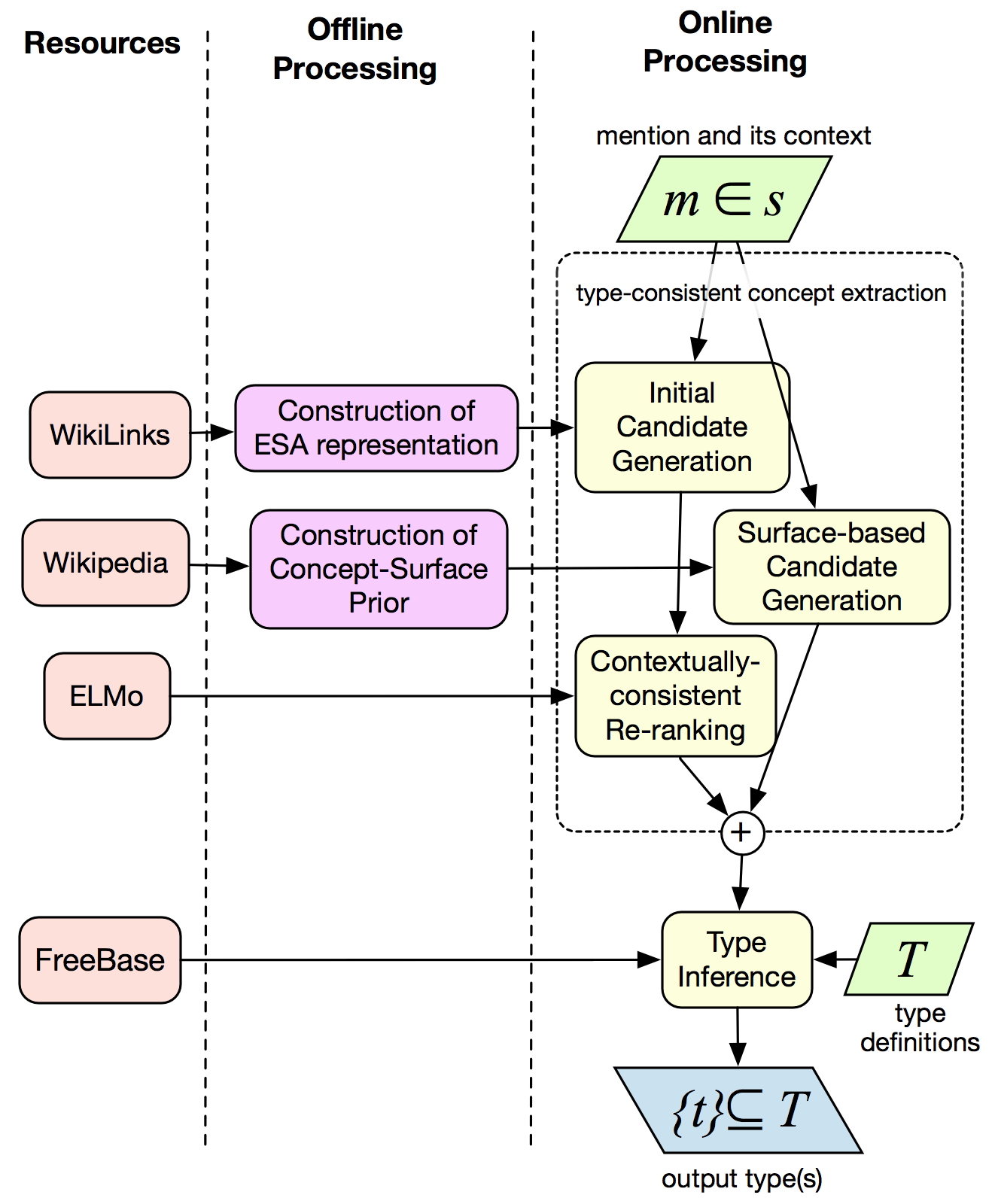}
    \caption{A high-level view of our approach. The \colorbox{lightgreen}{inputs} to the system are a mention $m$ in a context $s$, and type definitions $T$. The \colorbox{lightblue2}{output} is set of types $\{t\}$ in the type definition. The figure also highlights the input \colorbox{lightred2}{resources}, as well as  \colorbox{lightpurple}{offline} and \colorbox{lightyellow}{online processes}. 
    }
    \label{fig:diragram}
\end{figure}
Figure~\ref{fig:diragram} shows the high-level architecture of our proposed system. The inputs to the system are a mention $m$ in a sentence $s$, and a type definition $T$. The output of the system is a set of types $\{t_{\text{Target}}\} \subseteq T$ in the target taxonomy that best represents the given mention. 
The type definitions characterize the target entity-type space. In our experiments, we choose to use \freebase\ types to define the types across 7 datasets; that is, $T$ is a mapping from the set of \freebase\ types to the set of target types: $T: \{ t_{\text{FB}} \} \rightarrow  \{ t_{\text{Target}} \} $. 
This definition comprises many atomic definitions; for example, we can define the type {\tt \small location} as the disjunction of \freebase\ types like {\tt \small FB.location} and {\tt \small FB.geography}:
\resizebox{\linewidth}{!}{
\includegraphics{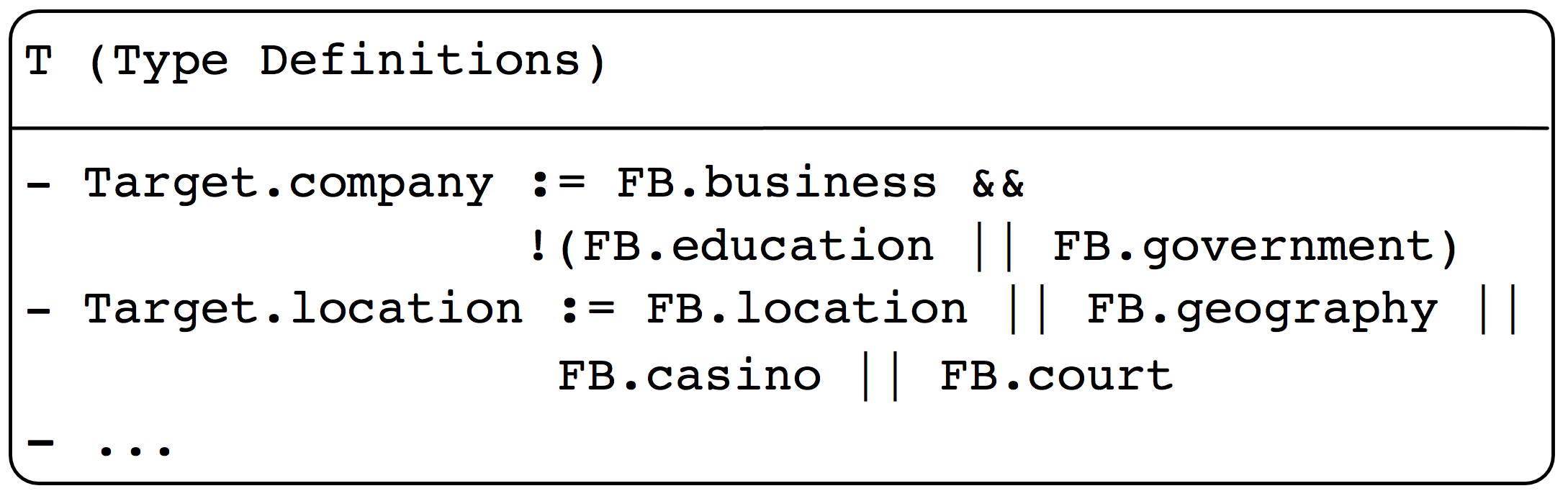}
}

The type definitions of a dataset reflect the understanding of a domain expert and the assumptions made in dataset design. Such definitions are often much cheaper to define, than to annotate full-fledged supervised datasets.
It is important to emphasize that, to use our system on different datasets, one does not need to retrain it; 
there is one single system used across different datasets, working with different type definitions. 

For notational simplicity, we define a few conventions for the rest of the paper. 
The notation $t \in T$, simply means $t$ is a member of the image of the map $T$ (i.e., $t$ is a member of the target types). For a fixed concept $c$, the notation $T(c)$ is the application of $T(.)$ on the \freebase\ types attached to the concept $c$. For a collection of concepts $C$, $T(C)$ is defined as $\bigcup_{c \in C} T(c)$. 
We use 
$T_{\text{coarse}}(.)$ to refer to the subset of coarse types of $T(.)$, while $T_{\text{fine}}(.)$ defines the fine type subset. 

Components in Figure~\ref{fig:diragram} are described in the following sections.

\begin{figure*}
    \centering
    \includegraphics[trim=0.2cm 1.1cm 0cm 0.4cm, scale=0.32]{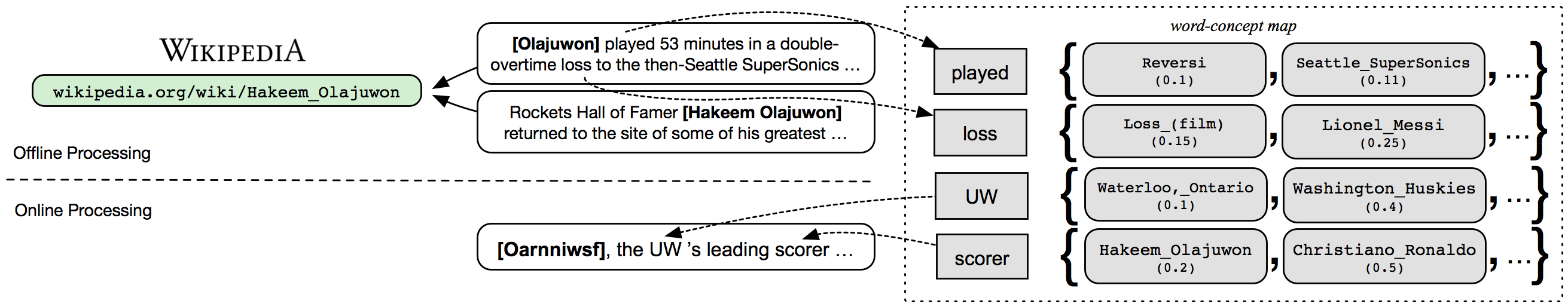}
    \caption{Extraction of topically relevant concepts. Word-concept map is pre-computed using \wikilinks\ and used to retrieve the most relevant concepts for a given mention (see \S\ref{subsec:step1:esa}).
    }
    \vspace{-0.15cm}
    \label{fig:step1:type-relevance}
\end{figure*}

\subsection{Initial Concept Candidate Generation} 
\label{subsec:step1:esa}
Given a mention, the goal of this step is to quickly generate a set of Wikipedia entries based on other words in the sentence. Since there are millions of entries in Wikipedia, it is extremely inefficient to go through all entries for each mention. We adopt ideas from explicit semantic analysis (ESA)~\cite{gabrilovich2007computing}, an approach to representing words with a vector of Wikipedia \emph{concepts}, and to providing fast retrieval of the relevant Wikipedia concepts via inverted indexing. 

\newcommand{\sent}{{\footnotesize \textsf{sent}}} 
\newcommand{\score}{{\footnotesize \textsf{score}}} 

In our construction we use the \wikilinks~\cite{singh2012wikilinks} corpus, which contains a total of 40 million mentions over 3 million concepts. Each mention in \wikilinks\ is associated with a Wikipedia concept. To characterize it formally, in the \wikilinks\ corpus, for each concept $c$, there are example sentences $\sent(c)=\{s_i\}$.

\paragraph{Offline computation:}
The first step is to construct an ESA representation for each word in the \wikilinks\ corpus. 
We create a mapping from each word in the corpus to the relevant concepts associated with it. 
The result is a map $\mathcal{S}$ from tokens to concepts: $\mathcal{S}: w \rightarrow \{ c, \score(c|w) \}$ (see Figure~\ref{fig:step1:type-relevance}), where $\score(c|w)$ denotes the association of the word $w$ with concept $c$, calculated as the sum of the TF-IDF values of the word $w$ in the sentences describing $c$:  
$$
\score(c|w) \triangleq \sum_{s \in \sent(c)}\sum_{w \in s} \text{tf-idf}(w,s).
$$
That is, we treat each sentence as a document and compute TF-IDF scores for the words in it. 

\paragraph{Online computation:}
For a given mention $m$ and its sentence context $s$, we use our offline word-concept map $\mathcal{S}$ to find the concepts associated with each word, and aggregate them to create a single list of weighted concepts; i.e., $\sum_{w \in s} \mathcal{S}(w)$. The resulting concepts are sorted by the corresponding weights, and the top $\ell_{ESA}$ candidates form a set $C_{\text{ESA}}$ which is passed to the next step.

\newcommand{\sentrep}[1]{{\footnotesize\textsf{SentRep}}(#1)}
\newcommand{\conceptrep}[1]{{\footnotesize\textsf{ConceptRep}}(#1)}
\newcommand{\consistency}{{\footnotesize \textsf{Consistency}}} 

\subsection{Context-Consistent Re-Ranking}
\label{subsec:step2:elmo}

After quick retrieval of the initial concept candidates, we re-rank concepts in $C_{\text{ESA}}$ based on context consistency between the input mention and concept mentions in \wikilinks. 

For this step, assume we have a representation that encodes the sentential information anchored on the mention. We denote this mention-aware context representation as \sentrep{$s| m$}. We define a measure of consistency between a concept $c$ and a mention $m$ in a sentence $s$: 
{
\begin{multline}
\label{eq:consistency}
\consistency(c, s, m) = \\ 
\hspace{0.6cm} \text{cosine}( \sentrep{s| m}, \conceptrep{c} ),
\end{multline}
}
\noindent where $\conceptrep{c}$ 
is representation of a concept: 
{
\small
\begin{multline*}
\small
\conceptrep{c} \triangleq \\ 
\hspace{0.7cm} \text{avg}_{\substack{s}} \left( \sentrep{s|c} \Big| s \in \wikilinks, c \in s) \right),
\end{multline*}
}
\noindent which is the average vector of the representation of all the sentences in \wikilinks\ that describe the given concept. 

We use pre-trained \elmo~\cite{peters2018deep}, a state-of-the-art contextual and mention-aware word representation. In order to generate $\sentrep{s| m}$, we run \elmo\ on sentence $s$, where the tokens of the mention $m$ are concatenated with ``\_'', and retrieve its \elmo\ vector 
as $\sentrep{s| m}$.

According to the consistency measure, we select the top $\ell_{\elmo}$ concepts for each mention. We call this set of concepts $C_{\elmo}$.

\subsection{Surface-Based Concept Generation}
\label{subsec:step3:prior}

While context often is a key signal for typing, one should not ignore the information included in the surface form of the mentions. If the corresponding concept or entity exists in Wikipedia, many mentions can be accurately grounded with only trivial prior probability $\mathbf{Pr}(\text{concept}|\text{surface})$. The prior distribution is pre-computed by calculating the frequency of the times a certain surface string refers to different concepts within Wikipedia. 

In the test time, for a given mention, we use the pre-computed probability distribution to obtain the most likely concept, $c_{\text{surf}} = \arg\max_c \mathbf{Pr}(c|m) $, for the given mention $m$. 


\newcommand{\countesa}[1]{{\footnotesize\textsf$\text{Count}_{ESA}$}(#1)}
\newcommand{\countelmo}[1]{{\footnotesize\textsf$\text{Count}_{ELMo}$}(#1)}
\newcommand{\counta}{\footnotesize\textsf{Count}}

\subsection{Type Inference}
\label{subsec:step3:type-inf}
Our inference algorithm starts with selection of concepts, followed by inference of coarse and fine types. Our approach is outlined in Algorithm~\ref{algorithm:inference} and explained below. 

\paragraph{Concept inference.}
To integrate surface-based and context-based concepts, we follow a simple rule: if the prior probability of the surface-based concept ($c_{\text{surf}}$) has confidence below a threshold $\lambda$, we ignore it; otherwise we include it among the concepts selected from context ($C_{\elmo}$), and only choose coarse and fine types from $c_{\text{surf}}$.

To map the selected concepts to the target entity types, we retrieve the \freebase-types of each concept and then apply the type definition $T$ (defined just before \S\ref{subsec:step1:esa}). 
In Algorithm~\ref{algorithm:inference}, the set of target types of a concept $c$ is denoted as $T(c)$. 
This is followed by an aggregation step for selection of a coarse type $t_{\text{coarse}} \in T_{\text{coarse}}(.)$, and ends with the selection of a set of fine types $\{t_{\text{fine}}\} \subseteq T_{\text{fine}}(.)$. 

\newcommand{\argminE}{\mathop{\mathrm{argmin}}}          
\newcommand{\argmaxE}{\mathop{\mathrm{argmax}}} 
\newcommand{\selectCoarse}{\footnotesize \textsf{SelectCoarse}}

\paragraph{Coarse type inference.}
Our type inference algorithm works in a relatively simple confidence analysis procedure. 
To this end, we define $\counta(t; C)$ to be the number of occurrences of type $t$ in the collection of concepts $C$: 
$$\counta(t; C) :=  |\{c: c \in C \text{ and }  t \in T(c) \} |.$$

In theory, for a sensible type $t$, the count of context-consistent concepts that have this type should be higher than that of the initial concept candidates. In other words, $\frac{\counta(t; C_\elmo) / \ell_{\elmo}}{ \counta(t; C_\esa) / \ell_\esa}   > 1$. We select the first concept (in the $C_{\elmo}$ ranking) which has some coarse type that matches this criterion. 
If there is no such concept, we use the coarse types of the highest scoring concept.
To select one of the coarse types of the selected concept, we let each concept of $C_\elmo$  vote based on its consistency score. We name this voting-based procedure $\selectCoarse(c),$ which selects one coarse type from a given concept: 
\begin{multline*}
\selectCoarse(c) \triangleq \\ 
\hspace{0.5cm} \argmaxE_{t} \sum_{\tilde{c} \in C_{\elmo}}
\sum_{t \in T_{\text{coarse}}(\tilde{c})} \consistency(\tilde{c}, s, m), 
\end{multline*}
where consistency is defined in Equation~(\ref{eq:consistency}). 


\paragraph{Fine type inference.}
With the selected coarse type, we take only the fine types that are compatible with the selected coarse type (e.g., the fine type {\tt \small /people/athlete} and the coarse type {\tt \small /people} are compatible). 

Among the compatible fine types, we further filter the ones that have better support from the context. Therefore, we select the fine types $t_f$ such that $\frac{\counta(t_f; C_\elmo)}{\counta(t_c; C_\elmo)} \geq \eta$, where $t_c$ is the previously selected coarse type which is compatible with $t_f$. 
Intuitively, the fraction filters out the fine-grained candidate types that don't have enough support compared to the selected coarse type. 


\begin{algorithm}[t!]
\sffamily
\SetNoFillComment
\small
\SetAlgoLined
\SetKwInOut{Input}{Input}
\SetKwInOut{Output}{Output}
\textbf{Input} {mention $m$ in sentence $s$, retrieved concepts $C_{\text{ESA}}, C_{\elmo}, c_{\text{surf}}$, and type definition $T$} \\ 
\textbf{Output} Inferred types $t_{\text{coarse}}$ 
and $\{t_{\text{fine}}\}.$ \\ 
\textbf{Define},
$r(t, t'; C, C') :=  \frac{ \counta(t; C) / |C| }{  \counta(t'; C') / |C'| }$,
$r(t; C, C') := r(t, t; C, C')$,
$r(t, t'; C,) := r(t, t'; C, C)$.\\ 
  \BlankLine

$\tau_{\text{surf}} \leftarrow  \{ t |  t \in T_{\text{coarse}}(c_{\text{surf}}), r(t; C_{\elmo}, C_{\text{ESA}}) > 1 \} $ \\ 
\eIf{ $\textbf{Pr}(c_{\text{surf}} | m) \geq \lambda$ {\bf and} $\tau_{\text{surf}} \neq  \emptyset$ 
} { 
    $t_{\text{coarse}} \leftarrow \text{SelectCoarse}(c_{\text{surf}})$ \\
    $\tilde{C} \leftarrow \{c_{\text{surf}}\} \cup C_{\elmo} $ \\
    $\{t_{\text{fine}} \} \leftarrow
    \left\{ 
    t_f \ \middle\vert \begin{array}{l}
    t_f \in T_{\text{fine}}(c_{\text{surf}})\text{,}\\ \text{compatible w/ } t_{\text{coarse}} \text{ and, } \\ 
    r(t_f, t_{\text{coarse}};  \tilde{C}) \geq \eta_s
    \end{array}
    \right\}
    $ \\ 
}{
$\tilde{C}_{\elmo} \leftarrow   
\left\{ c \ \middle\vert \begin{array}{l}
    c \in C_{\elmo}, \exists t \in T_{\text{coarse}}(c) \\
    r(t; C_{\elmo}, C_{\text{ESA}}) > 1
  \end{array}\right\} $\\ 
    
    
    \eIf{
    $\tilde{C}_{\elmo} = \o$
    }{
        $\tilde{c} \leftarrow \argmaxE_{c \in C_{\elmo} } \text{Consistency}(c,s,m)$ \\ 
    } {
        $\tilde{c} \leftarrow \argmaxE_{c \in \tilde{C}_{\elmo} } \text{Consistency}(c,s,m)$ \\ 
    }
    
    $t_{\text{coarse}} \leftarrow \text{SelectCoarse}(\tilde{c}) $ \\
    $\{t_{\text{fine}} \} \leftarrow 
    \left\{ 
    t_f \ \middle\vert \begin{array}{l}
    t_f \in T_{\text{fine}}(C_{\elmo}) \text{,} \\ 
    \text{compatible w/ } t_{\text{coarse}}  \text{ and,} \\  r(t_f, t_{\text{coarse}}; C_{\elmo}) \geq \eta_c
    \end{array}
    \right\}
    $\\ 
}
 \caption{Type inference algorithm}
 \label{algorithm:inference}
\end{algorithm}
\vspace{-0.0cm}

\newcommand{\elmonn}{\textsc{{ELMoNN}}}
\newcommand{\wikiBaseline}{\textsc{{WikifierTyper}}}


\begin{table*}[thp]
\centering
\small 
\resizebox{15.9cm}{!}{
\begin{tabular}{cC{5.7cm}cccccccccc}
    \toprule
     & & & \multicolumn{3}{c}{\figer} &  \multicolumn{3}{c}{\bbn} & \multicolumn{3}{c}{\ontonotesFine}   \\ 
     \midrule
      & Approach & Trained on & \accuracy & \fOneMa & \fOneMi & \accuracy & \fOneMa & \fOneMi & \accuracy & \fOneMa & \fOneMi  \\ 
     \cmidrule(lr){2-2} \cmidrule(lr){3-3} \cmidrule(lr){4-6} \cmidrule(lr){7-9} \cmidrule(lr){10-12} 
    \parbox[t]{1mm}{\multirow{8}{*}{\rotatebox[origin=c]{90}{\small  Closed Type}}} & \afet*~\cite{Ren2016AFETAF}   
    & \figer & 53.3 & 69.3 & 66.4 & - & - & - & - & - & - \\ 
     & \nfetc*\cite{xu2018neural} & \figer  &  \underline{68.9}	& \underline{81.9}	& \underline{79.0} & - & - & - & - & - & -  \\ 
     \cdashlinelr{2-12}
     & \afet*~\cite{Ren2016AFETAF} & \bbn & - & - & - &   68.3 & 74.4  &  74.7  & - & - & - \\ 
     & \abishek*~\cite{abhishek2017fineGrained} & \bbn & - & - & - & \underline{73.3} & \underline{79.1} & \underline{79.2} & - & - & - \\  
     \cdashlinelr{2-12}
    &  \afet*~\cite{Ren2016AFETAF} & \ontonotesFine & - & - & - & - & - & - & 55.1 & 71.1 & 64.7  \\
    &  \nfetc*~\cite{xu2018neural} & \ontonotesFine &  - & - & - & - & - & - & \underline{60.2} & \underline{76.4} & \underline{70.2} \\
\midrule
    \parbox[t]{1mm}{\multirow{6}{*}{\rotatebox[origin=c]{90}{\small Open Type}}} & \multicolumn{1}{c}{\multirow{3}{3.4cm}{\hspace{0.85cm}\otyper\hspace{0.85cm}\cite{yuan2018otyper}}}   &  \figer &  \cellcolor{graaay} 47.2 & \cellcolor{graaay} 69.1 & \cellcolor{graaay} 67.2 & 27 & 50.3 & 49.5 & 31.6 & 34.5 & 32.1  \\
    &   &  \bbn & 5.3 & 11.5 & 11.5 & \cellcolor{graaay} 29 & \cellcolor{graaay} 54.4 &  \cellcolor{graaay} 48.8 & 2.5 & 5.1 & 5.4  \\
    &   &  \ontonotesFine & 0.4 & 15.6 & 16.8 & 23.6 & 51.1 & 47.9 & \cellcolor{graaay} 31.8 & \cellcolor{graaay} 39.1 & \cellcolor{graaay} 36  \\
    \cdashlinelr{2-12}
    & \elmonn       & $\times$ & 21.5   & 57.7	& 53.8  & 	49.3	& 	68.4	& 	66.2 & 0.5 &  21.2 & 21.8  \\ 
    & \wikiBaseline & $\times$ & 17.2  & 33.3	& 46.2  & 	45.8 & 52.3 &  66.1 & 47.8 &	65.6 &	58.2 \\ 
    & \ours\ (ours) & $\times$ &  \bf 58.8 & \bf  74.8	& \bf  71.3  &   \bf  61.8 & \bf  74.6 &  \bf  74.9 & \bf  50.7 & \bf  66.9 &  \bf  60.8 \\
    \bottomrule
\end{tabular}
}
\vspace{-0.15cm}
\caption{
\small 
Evaluation of fine-grained entity-typing: we compare our system with state-of-the-art systems (\S\ref{sec:fine:grained:exp})
For each column, the best zero-shot and overall results are {\bf bold-faced} and \underline{underlined}, respectively. Numbers are $F1$ in percentage. For supervised systems, we report their in-domain performances, since they do not transfer to other datasets with different labels. 
For \otyper, cells with \colorbox{graaay}{gray} color indicate \emph{in-domain} evaluation, which is the setting in which it has the best performance.
Our system outperforms all the other zero-shot baselines, and achieves competitive results compared to the best supervised systems. 
}
\label{tab:figer-results}
\end{table*}

\begin{table*}
    \small
    \centering
    \resizebox{13.5cm}{!}{
    \begin{tabular}{ccccccccccc}
        \toprule
            & & \multicolumn{3}{c}{\ontonotes} & \multicolumn{3}{c}{ \conll } & \multicolumn{3}{c}{ \muc } \\
        \midrule
           System & Trained on & PER & LOC & ORG & PER & LOC & ORG & PER & LOC & ORG \\
        \cmidrule(lr){1-1} \cmidrule(lr){2-2} \cmidrule(lr){3-5} \cmidrule(lr){6-8} \cmidrule(lr){9-11}
        \cogcompnlp & \ontonotes  & \cellcolor{graaay} \underline{98.4} & \cellcolor{graaay} \underline{91.9} & \cellcolor{graaay} \underline {97.7} &  83.7 & 70.1 & 68.3 & 82.5 & 76.9 & 86.7 \\
        \cogcompnlp & \conll & \bf{94.4} & 59.1 & \bf{87.8} & \cellcolor{graaay} \underline{95.6} & \cellcolor{graaay} \underline {92.9} & \cellcolor{graaay} \underline{90.5} &  \bf \underline{90.8} & 90.8 & 90.9 \\
        \ours~(ours) & $\times$ & 88.4 & \bf{70.0} & 85.6 & \bf{90.1} &\bf{80.1} & \bf{73.9} & 87.8 & \bf \underline{90.9} & \bf \underline{91.2} \\
        \bottomrule 
    \end{tabular}
    }
    \vspace{-0.15cm}
    \caption{
    \small 
    Evaluation of coarse entity-typing (\S\ref{sec:coarse:type:exp}): we compare two supervised entity-typers with our system. For the supervised systems, cells with \colorbox{graaay}{gray} color indicate \emph{in-domain} evaluation. For each column, the best, out-of-domain and overall results are {\bf bold-faced} and \underline{underlined}, respectively. Numbers are $F1$ in percentage. In most of the out-of-domain settings our system outperforms the supervised system. 
    }
    \label{tab:coarse-ner}
\end{table*}

\section{Experiments}
\label{sec:experiments}
Empirically, we study the behavior of our system compared to published results. 
All the results are reproduced
except the ones indicated by $*,$ which are directly cited from their corresponding papers.
\paragraph{Datasets.}
In our experiments, we use a wide range of typing datasets: 
\begin{itemize}[leftmargin=*,noitemsep,nolistsep,topsep=0pt]
    \item For \textbf{coarse} entity typing, we use \muc~\cite{grishman1996message}, \conll~\cite{tjong2003introduction}, and \ontonotes~\cite{Hovy2006OntoNotesT9}. 
    \item For \textbf{fine} typing, we focus on \figer~\cite{ling2012fine}, \bbn~\cite{weischedel2005bbn}, and \ontonotesFine~\cite{Gillick2014ContextDependentFE}. 
    \item In addition to the news NER, 
we use the \bionlp\ dataset~\cite{deleger2016overview}, with contain mentions of \emph{bacteria} or other notions, extracted from sentences of scientific papers. 
\end{itemize}

\paragraph{\ours's parameters.}
We use different type definitions for each dataset. 
In order to design type definitions for each dataset, we follow in the footsteps of~\namecite{abhishek2017fineGrained} and randomly sample $10\%$ of the test set. For the experiments,
we exclude the sampled set.
For completeness, we have included the type definitions of the major experiments in Appendix~\ref{sec:supp3}.  

The parameters are set universally across different experiments. For parameters that determine the number of extracted concepts, we use $\ell_{\text{ESA}} = 300$ and $\ell_{\elmo} = 20$, which are based on the upper-bound analysis in Appendix~\ref{sec:supp1}.  
For other parameters, we set  $\lambda = 0.5$, $\eta_s = 0.8$ and $\eta_c = 0.3$, based on the \figer\ dev set. 
We emphasize that these parameters are universal across our evaluations. 

\paragraph{Evaluation metrics.}
\label{supp:per-type-eval}
Given a collection of mentions $M$, denote the set of gold types and predicted types of a mention $m \in M$ as $T_g(m)$ and $T_p(m)$ respectively. We define the following metrics for our evaluations: 
\begin{itemize}[leftmargin=*,noitemsep,nolistsep,topsep=0pt]
    \item Strict Accuracy (\accuracy): $\frac{|\{m | T_g(m) = T_p(m) \}|}{|M|}$. 
    \item Macro F1 (\fOneMa): Macro Precision is defined as $\frac{1}{|M|} \sum_{m \in M} \frac{|T_p(m) \cap T_g(m)| }{|T_p(m)|}$. With this, the definitions of Macro recall and F1 follow. 
    \item Micro F1 (\fOneMi): The precision is defined as $\frac{\sum_{m \in M} |T_p(m) \cap T_g(m)|}{ \sum_{m \in M} |T_p(m)| }$, and the Micro recall and F1 follow the same pattern. 
\end{itemize}

In the experiment in \S\ref{sec:exp1:unseen}, to evaluate systems on unseen types 
we 
used modified versions of metrics. 
Let $G(t)$ be the number of mentions with gold type $t$,  $P(t)$ be the number of mentions predicted to have type $t$, $C(t)$ be the number of mentions correctly predicted to have type $t$: 
\begin{itemize}[leftmargin=*,noitemsep,nolistsep,topsep=0pt]
    \item The precision corresponding to \fOneMat\ is defined as $\sum_{t} \frac{C(t)}{P(t)} \frac{G(t)}{\sum_{t^\prime} G(t^\prime)}$; recall follows the same pattern. 
    \item The precision corresponding to \fOneMit\ 
    is defined as $\frac{\sum_{t} C(t)}{\sum_{t}P(t)}$; recall follows the same pattern. 
\end{itemize}

\paragraph{Baselines.}
To add to the best published results on each dataset, we create two simple and effective baselines. The first baseline, \elmonn, selects the nearest neighbor types to a given mention, where \emph{mentions} and \emph{types} are represented by \elmo\ vectors. To create a representation for each type $t$, we average the representation of the \wikilinks\ sentences that contain mentions of type $t$ (as explained in \S\ref{subsec:step2:elmo}). 
Our other baseline, \wikiBaseline, uses Wikifier~\cite{tsai2016cross} to map the mention to a Wikipedia concept, followed by mapping to \freebase\ types, and finally projecting them to the target types, via type definition function $T(.)$. 
Additionally, to compare with published zero-shot systems, we compare our system to \otyper, a recently published \emph{open}-typing system. Unfortunately, to the best of our knowledge, the systems proposed by \namecite{ma2016label,huang2016building} are not available online for empirical comparison. 


\subsection{Fine-Grained Entity Typing }
\label{sec:fine:grained:exp}
We evaluate our system for \emph{fine-grained} named-entity typing. 
Table~\ref{tab:figer-results} shows the evaluation result for three datasets, \figer, \bbn, and \ontonotesFine. We report our system's performance, our zero-shot baselines, and two supervised systems (\afet, plus the-state-of-the-art), for each dataset. 
There is no easy way to transfer the supervised systems across datasets, hence no out-of-domain numbers for such systems.  
For each dataset, we train \otyper\ and evaluate on the test sets of all the three datasets. In order to run \otyper\ on different datasets, we disabled original dataset-specific entity and type features. 
As a result, among the \emph{open} typing systems, our system has significantly better results. In addition, our system has competitive scores compared to the supervised systems.

\begin{table}[t]
\centering
\small 
\resizebox{7.3cm}{!}{
\begin{tabular}{ccc}
\toprule
Approach & \fOneMat & \fOneMit \\
\cmidrule(lr){1-1} \cmidrule(lr){2-3}
 \elmonn   & 63.1 & 53.8\\ 
 \wikiBaseline & 53.0 & 43.9\\ 
 \otyper~\cite{yuan2018otyper}   &  50.6	&   23.4 \\ 
 \ours\ (ours)   &  {\bf 71.7 }	& {\bf  71.1 } \\ 
\bottomrule
\end{tabular}
}
\vspace{-0.15cm}
\caption{
    Comparing systems where no labels (types) are seen a priori (\S\ref{sec:exp1:unseen}). 
}
\label{table:unseen-results}
\end{table}

\begin{table}[t]
\centering
\small
\resizebox{7.3cm}{!}{
\begin{tabular}{cccc}
\toprule
    System & Bacteria & not-Bacteria & Overall \\  
    \cmidrule(lr){1-1} \cmidrule(lr){2-4}
    \wikiBaseline & 54.8 & \bf 86.2 & 70.5 \\
    \elmonn & 67.6  & 81.2    & 74.4   \\
    \ours\ (ours) &\bf  68.1 & 84.2 & \bf76.2 \\  \bottomrule
\end{tabular}
}
\vspace{-0.15cm}
\caption{
Results of the system classifying mentions to ``\emph{bacteria}'' or something else (\S\ref{sec:biology}). Numbers are $F1$ in percentage. 
}
\label{table:figer-results}
\end{table}

\begin{table*}
\centering
\small 
\resizebox{13.5cm}{!}{
\begin{tabular}{C{3.6cm}ccccccccc}
\toprule
      & \multicolumn{3}{c}{\figer} &  \multicolumn{3}{c}{\bbn} & \multicolumn{3}{c}{\ontonotesFine}   \\ 
\midrule 
    Approach &  \accuracy & \fOneMa & \fOneMi & \accuracy & \fOneMa & \fOneMi & \accuracy & \fOneMa & \fOneMi  \\
    \cmidrule(lr){1-1} \cmidrule(lr){2-4} \cmidrule(lr){5-7} \cmidrule(lr){8-10}
    \ours~(ours)  & 58.8 & 74.8	& 71.3  &   61.8 & 74.6 &  74.9 & 50.7 & 66.9 &  60.8 \\ 
    \cdashlinelr{1-10}
    no surface-based concepts  & -8.8	& -7.5  & -9.2 & -12.9 & -7.0 & -8.6  & -1.8	& -1.2	&  -0.1\\ 
    no context-based concepts  & -39.3  & -42.1 & -25.4 & -36.4 & -31.0 & -13.9 & -10.0 & -12.3 & -7.4\\ 
\bottomrule
\end{tabular}
}
\vspace{-0.15cm}
\caption{
\small
Ablation study of different ways in which concepts are generated in our system (\S\ref{supp:ablation}). The first row shows performance of our system on each dataset, followed by the \emph{change} in the performance upon dropping a component. 
While both signals are crucial, contextual information is playing more important role than the mention-surface signal. 
}
\label{tab:ablation}
\end{table*}

\subsection{Coarse Entity Typing}
\label{sec:coarse:type:exp}
In Table~\ref{tab:coarse-ner} we study entity typing for the coarse types on three datasets. We focus on three types that are shared among the datasets: \emph{PER}, \emph{LOC}, \emph{ORG}. 
In coarse-entity typing, the best available systems are heavily supervised. In this evaluation, we use gold mention spans; i.e., we force the decoding algorithm of the supervised systems to select the best of the three  classes for each gold mention. 
As expected, the supervised systems have strong in-domain performance. However, they suffer a significant drop when evaluated in a different domain. Our system, while not trained on any supervised data, achieves better or comparable performance compared to other supervised baselines in the out-of-domain evaluations. 

\subsection{Typing of Unseen Types within Domain}
\label{sec:exp1:unseen}
We compare the quality of \emph{open} typing, in which the target type(s) have not been seen before. 
We compare our system to \otyper, which relies on supervised data to create representations for each type; however, it is not restricted to the \emph{observed types}. 
We follow a similar setting to \namecite{yuan2018otyper} and split the \figer\ test in folds (one fold per type) and do cross-validations. For each fold, mentions of only one type are used for evaluation, 
and the rest are used for training \otyper. 
To be able to evaluate on \emph{unseen} types
(only for this experiment), we use modified metrics \fOneMat\ and \fOneMit\ that measure \emph{per type} quality of the system (\S\ref{supp:per-type-eval}). In this experiment, we focus on a within-domain setting, and show the results of transfer across genres in the next experiments. 
The results are summarized in Table~\ref{table:unseen-results}. We observe a significant margin between \ours\ and other systems, including \otyper.

\subsection{Biology Entity Typing}
\label{sec:biology}
We go beyond the scope of popular entity-typing tasks, and evaluate the quality of our system on a dataset that contains sentences from scientific papers~\cite{deleger2016overview}, which makes it different from other entity-typing datasets. 
The mentions refer to either ``\emph{bacteria}'', or some miscellaneous class (two class typing). 
As indicated in Table~\ref{table:figer-results}, our system's overall scores are higher than our baselines. 







\subsection{Ablation Study}
\label{supp:ablation}

We carry out ablation studies that quantify the contribution of surface information (\S\ref{subsec:step3:prior}) and context information (\S\ref{subsec:step2:elmo}). As Table~\ref{tab:ablation} shows, both factors are crucial and complementary for the system. However, the contextual information seems to have a bigger role overall. 

We complement our qualitative analysis with the quantitative share of each component. In 69.3\%, 54.6\%, and 69.7\% of mentions, our system uses the context information (and ignores the surface), in \figer, \bbn, and \ontonotesFine\ datasets, respectively, underscoring the importance of contextual information.  


\subsection{Error Analysis}
\label{sec:error:analyais}
We provide insights into specific reasons for the mistakes made by the system. 
For our analysis, we use the erroneous decisions in the \figer\ dev set. Two independent annotators label the cause(s) of the mistakes, resulting in $83\%$ agreement between the annotators. The disagreements are later reconciled by an adjudication step.  
\begin{enumerate}[leftmargin=*,noitemsep,nolistsep,topsep=0pt]
    \item \emph{Incompatible concept, due to context information:} Ambiguous contexts, or short ones, often contribute to the inaccurate mapping to concepts. In our manual annotations, 23.3\% of errors are caused, at least partly, by this issue. 
    \item \emph{Incompatible concept, due to surface information:}
        Although the prior probability is high, the surface-based concept could be wrong. About 26\% of the errors are partly due to the surface signal errors. 
    \item \emph{Incorrect type, due to type inference:} Even when the system is able to find several type-compatible concepts, it can fail due to inference errors. This could happen if the types attached to the type-compatible concepts are not the majority among other types attached to other concepts. This is the major reason behind 56.6\% of errors. 
    \item \emph{Incorrect type, due to type definition:} Some errors are caused by the inaccurate definition of the type mapping function $T$. About 23\% of the mistakes are partly caused by this issue. 
\end{enumerate}
Note that each mistake could be caused by multiple factors; in other words, the above categories are not mutually disjoint events. A slightly more detailed analysis is included in Appendix~\ref{sec:supp3:error:analaysis}.

\section{Conclusion}
\label{sec:conclusion}
Moving beyond a fully supervised paradigm and scaling entity-typing systems to support bigger type sets  is a crucial challenge for NLP. In this work, we have presented \ours, a zero-shot open entity typing framework.  The significance of this work is threefold. 
First, the proposed system does not require task-specific labeled data. Our system relies on type definitions, which are much cheaper to obtain than annotating 
thousands of examples. 
Second, our system outperforms existing state-of-the-art zero-shot systems by a significant margin. Third, we show that without reliance on task-specific supervision, one can achieve relatively robust transfer across datasets. 





\section{Acknowledgement}
The authors would like to thank Jennifer Sheffield, Stephen Mayhew, and Qiang Ning for invaluable suggestions. 
This work was supported by Contract HR0011-15-2-0025 with the US Defense Advanced Research Projects Agency (DARPA). Approved for Public Release, Distribution Unlimited. The views expressed are those of the authors and do not reflect the official policy or position of the Department of Defense or the U.S. Government.
Additionally, this research is supported by grants from Google and the Allen Institute for Artificial Intelligence (\url{allenai.org}). 
\bibliography{ccg,ref}
\bibliographystyle{acl_natbib_nourl}

\clearpage

\appendix


\begin{center}
{\Large \textbf{Supplemental Material for: ``Zero-Shot Open Entity Typing as Type-Compatible Grounding"}}
\end{center}

\section{Coverage Analysis for Concept Selection}
\label{sec:supp1}
To better understand the behavior of our concept retrieval, we perform an upper-bound analysis. Assuming that our type inference given concepts is perfect (an oracle inference), we want to estimate the best achievable score (upper-bound analysis). We perform this analysis in the output of two stages in the system, 
\begin{enumerate}[leftmargin=*,noitemsep,nolistsep,topsep=0pt]
    \item Step 1: After initial concepts extraction (\S\ref{subsec:step1:esa}). 
    \item Step 2: After \elmo\ reranking   (\S\ref{subsec:step2:elmo}). 
\end{enumerate}

Figure~\ref{fig:upper-bound-analysis} shows a summary of the upper-bound curves  for the output of the Step 1 and the Step 2. 
From these analysis we set the parameters $\ell_{\esa}$ and $\ell_{\elmo}$.
From the blue curve it is evident that after about 200 concepts, there is almost high-coverage. With this, we set the parameter $\ell_{\esa} = 300$. Furthermore, the red curve shows the strong coverage, even with a few dozen candidates. In our experiments, we choose top $\ell_{\elmo}=20$ concepts in the output of Step 2. 

\definecolor{lightblue}{RGB}{110, 150, 255}
\definecolor{lightred}{RGB}{255, 150, 150}

\begin{figure}[h]
    \centering
    \includegraphics[trim=2.3cm 0.5cm 0cm 0cm, clip=false, scale=0.26]{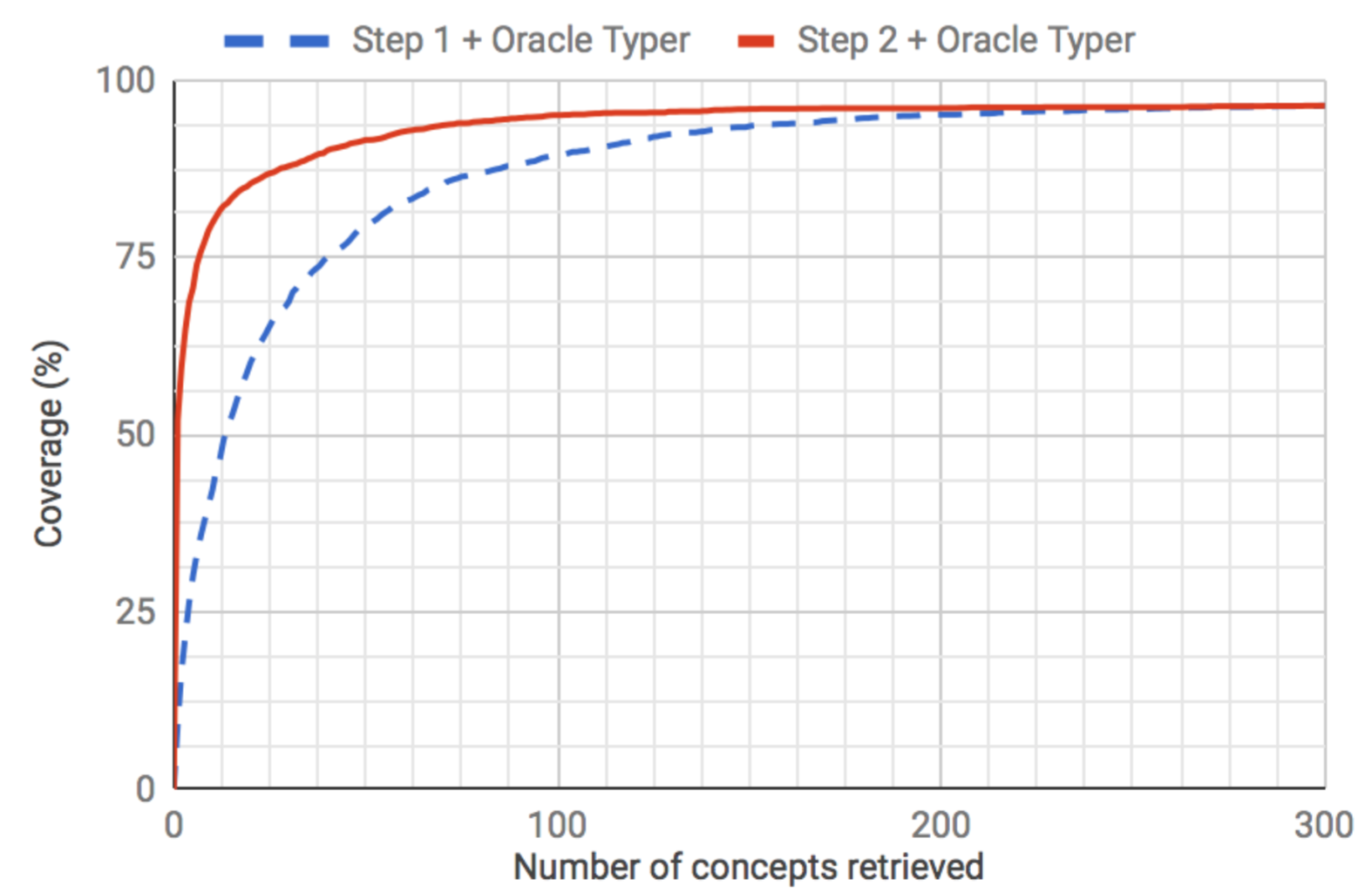}
    \caption{Upper-bound analysis: The y-axis shows the percentage of the instances with their types covered by the retrieved concepts. The x-axis shows the number of concepts retrieved, with a) \colorbox{lightblue}{Step 1} (dashed), b) \colorbox{lightred}{Step 2} (solid). }
    \label{fig:upper-bound-analysis}
\end{figure}

\section{Output for Each Step: An Example}
\label{sec:supp2}
\begin{figure*}[h]
    \centering
    \includegraphics[trim=3cm 0cm 0cm 0cm, clip=false, scale=0.41]{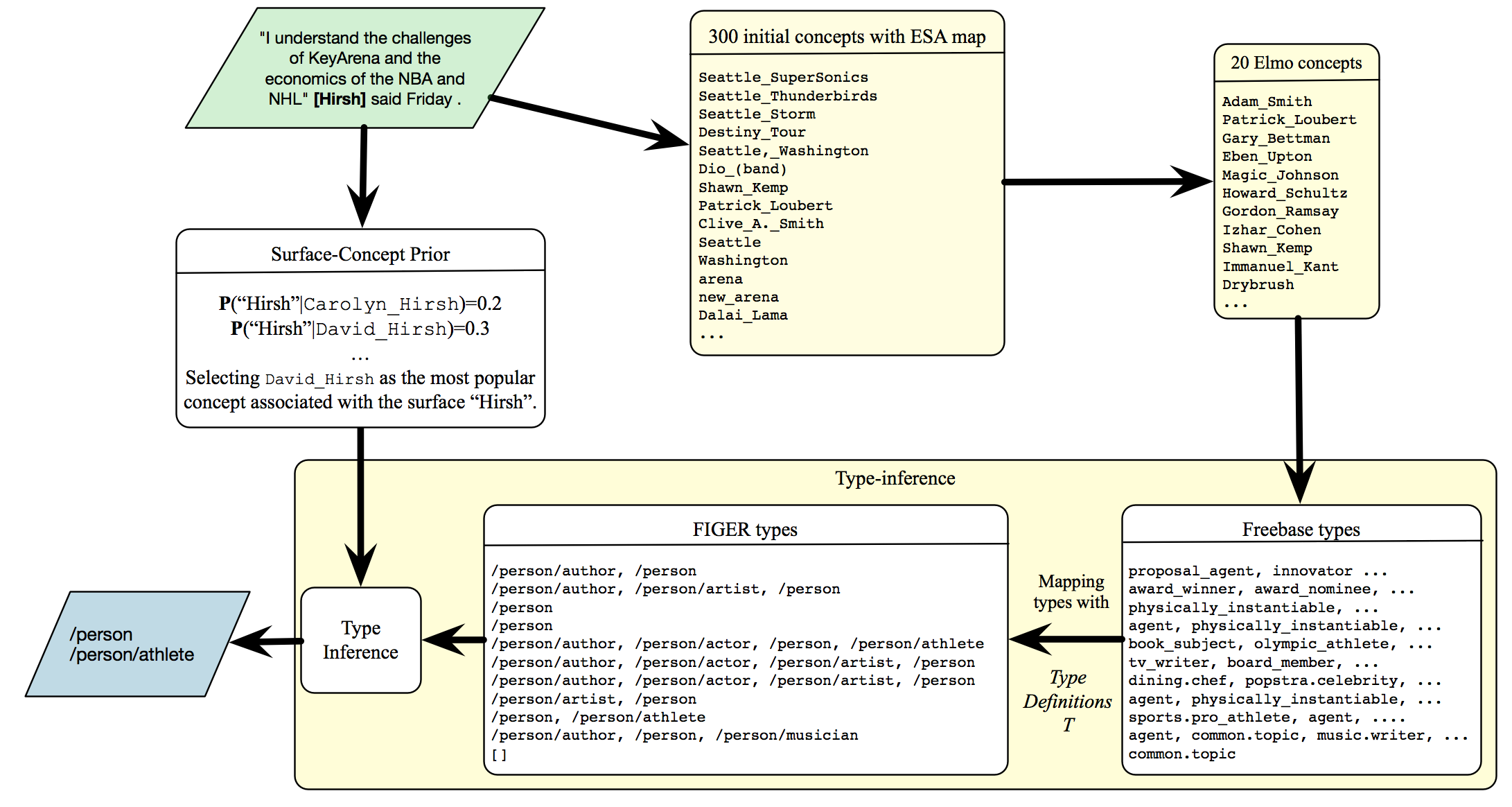}
    \caption{The output of each step in our system.}
    \label{fig:example}
\end{figure*}

To give a better intuition about every single step in our algorithm, we plot the outputs of each step in our construction in Figure~\ref{fig:example}. The details of how each step is done is included in \S\ref{sec:model}. 

\section{Error Analysis}
\label{sec:supp3:error:analaysis}
We expand a little bit on the errors analysis introduced in \S\ref{sec:error:analyais}. 
The first two categories consist of the scenarios when context and mention-surface information result in a type-compatible concept (hence incorrect type). The last two categories are due to inaccurate  mapping of (\emph{mostly} type-compatible) concepts to a collection of types.  
\begin{enumerate}
    \item \emph{Inconsistent concept, due to context information:} A short and ambiguous context could result in noisy concepts. In the following example, the majority of the selected concepts based on context are of type {\tt \small /politician}, however the correct label is {\tt \small /event}: \\ 
    {
        \small
        \emph{[The Fellows Forum], concerned in part with the induction of newly elected fellows, is just one event of the association’s annual meeting.}
    }
        
    \item \emph{Inconsistent concept, due to surface information:} When a mention used in a sense other than its most-popular sense, it could result in mistakes. 
    In the following example, while from the context it is clear that ``Utah" is a {\tt \small sports\_team}, the surface-string has a stronger association with {\tt \small Colorado}, which incorrectly results in the type {\tt \small /location}:\\ 
    {
    \small 
    \emph{The biggest cause for concern for McGuff is the bruised hamstring Regina Rogers suffered against [Utah] last Saturday.}
    }
    \item \emph{Incorrect type, due to type inference:} even when the system is able to find  type-compatible concepts, it still fails to infer the correct type, if the types attached to the type-compatible are the majority among other types. In the following example, while there are concepts of type {\tt \small /person}, the overall decision is incorrectly dominated by other concepts of type {\tt \small /location}: \\ 
    {
    \small 
    \emph{After years of wanting to curate such an exhibition, Wieczorek has collaborated with the Henry Art Gallery to feature 26 pieces of [Balth]'s ``Videowatercolors."}
    }\\ 
    The following is another example where the system fails to find the correct fine-type {\tt \small /person/musician}, since the overall decision is dominated by other concepts with type {\tt \small /person}, but not a musician (e.g. {\tt \small Blair\_Waldorf}).  \\ 
    {
    \small 
    \emph{``That's what I love about [Balth]'s art."}
    }
    
    \item \emph{Incorrect type, due to type definition:} Some errors are caused by inaccurate definition for the type mapping function $T$. 
    the following example, the mention gets mapped to an approximately correct concept {\tt infantry }, but the system  fails to map it to the correct type due to the limitations of the type definitions. 
    \\
    {
    \small 
    \emph{``When he left the [Army], Spencer got a job in Bozeman, where he used acupuncture to save a dog that couldn't walk anymore."}
    }
    
\end{enumerate}

\section{Type Definitions}
\label{sec:supp3}

We include the type definitions used in the experiments here for completeness. The types on the left are for the target dataset, and the types on the right, are the types from \freebase, combined with logical operators AND ($\with \with$), OR ($||$), NOT (!), and any-type wildcard character (*).

For the type definitions of the \figer\ dataset, we use the rules provided by (Ling and Weld, 2012), in addition to a few more definitions (Listing 1).

\begin{lstlisting}[caption={Additional definition used for FIGER beyond given mapping}]
/ORGANIZATION := /organization/organization
/LOCATION := /LOCATION || /BUILDING || /transportation/road
/ORGANIZATION/COMPANY := (/ORGANIZATION/COMPANY || /NEWS_AGENCY) $\with \with$ !(/ORGANIZATION/EDUCATIONAL_INSTITUTION || /ORGANIZATION/SPORTS_LEAGUE)
/WRITTEN_WORK := /WRITTEN_WORK $\with \with$ !/NEWS_AGENCY
/ART := /ART || /WRITTEN_WORK
\end{lstlisting}

\begin{lstlisting}[caption={Type definition used of the \bbn\ set.}]
/ORGANIZATION := /ORGANIZATION/ORGANIZATION
/PERSON := /PEOPLE/PERSON
/PLANT := /BASE/PLANTS/PLANT
/BUILDING := /ARCHITECTURE/BUILDING
/DISEASE := /MEDICINE/DISEASE
/LANGUAGE := /LANGUAGE/HUMAN_LANGUAGE
/LAW := /LAW $\with \with$ !/ORGANIZATION
/ANIMAL := /BIOLOGY/ANIMAL
/GPE/CITY := /LOCATION/CITYTOWN
/GPE/COUNTRY := /LOCATION/COUNTRY
/GPE/STATE_PROVINCE := /LOCATION/CN_PROVINCE || /BASE/LOCATIONS/STATES_AND_PROVENCES
/LOCATION := /LOCATION/LOCATION
/LOCATION/CONTINENT := /LOCATION/CONTINENT || /BASE/LOCATIONS/CONTINENTS 
/LOCATION/RIVER := /GEOGRAPHY/RIVER
/LOCATION/LAKE_SEA_OCEAN := /GEOGRAPHY/BODY_OF_WATER $\with \with$ !/LOCATION/RIVER
/LOCATION/REGION := /LOCATION/STATISTICAL_REGION
/FAC/AIRPORT := /AVIATION/AIRPORT
/FAC/HIGHWAY_STREET := /TRANSPORTATION/ROAD
/FAC/BRIDGE := /TRANSPORTATION/BRIDGE
/GAME := /CVG/COMPUTER_VIDEOGAME
/PRODUCT/VEHICLE := /AUTOMOTIVE/MODEL || /AVIATION/AIRCRAFT_MODEL
/PRODUCT/WEAPON := /LAW/INVENTION
/WORK_OF_ART/BOOK := /BOOK/WRITTEN_WORK
/WORK_OF_ART/SONG := /MUSIC/COMPOSITION
/WORK_OF_ART/PAINTING := /VISUAL_ART/ARTWORK
/WORK_OF_ART/PLAY := /THEATER/PLAY
/EVENT := /TIME/EVENT
/EVENT/WAR := /MILITARY/WAR
/EVENT/HURRICANE := /METEOROLOGY/TROPICAL_CYCLONE
/SUBSTANCE/FOOD := /FOOD/FOOD
/SUBSTANCE/DRUG := /MEDICINE/DRUG
/SUBSTANCE/CHEMICAL := /CHEMISTRY/CHEMICAL_COMPOUND
/ORGANIZATION/HOTEL := /TRAVEL/HOTEL
/ORGANIZATION/HOSPITAL := /MEDICINE/HOSPITAL
/ORGANIZATION/CORPORATION := /BUSINESS/EMPLOYER $\with \with$ !/ORGANIZATION/GOVERNMENT
/ORGANIZATION/POLITICAL := /GOVERNMENT/POLITICAL_PARTY
/ORGANIZATION/RELIGIOUS := /RELIGION/RELIGION $\with \with$ !/LOCATION/CONTINENT
/ORGANIZATION/EDUCATIONAL := /EDUCATION/ACADEMIC_INSTITUTION || /EDUCATION/EDUCATIONAL_INSTITUTION
/ORGANIZATION/GOVERNMENT := /GOVERNMENT/GOVERNMENT_AGENCY || /GOVERNMENT/GOVERNMENTAL_BODY || /GOVERNMENT/GOVERNMENT 
\end{lstlisting}

\begin{lstlisting}[caption={Type definition used of the \ontonotesFine\ set.}]
/PERSON := /PEOPLE/PERSON
/PERSON/ARTIST/AUTHOR := /BOOK/AUTHOR
/PERSON/ARTIST/ACTOR := /FILM/ACTOR
/PERSON/ARTIST/MUSIC := /MUSIC/ARTIST	
/PERSON/ATHLETE	:= /SPORTS/PRO_ATHLETE	
/PERSON/DOCTOR	/MEDICINE/PHYSICIAN	
/PERSON/POLITICAL_FIGURE := /GOVERNMENT/POLITICIAN	
/PERSON/LEGAL := /BASE/CRIME/CRIMINAL_DEFENCE_ATTORNEY || /BASE/CRIME/LAWYER_TYPE || /LAW/JUDGE
/PERSON/TITLE := /FICTIONAL_UNIVERSE/FICTIONAL_JOB_TITLE || /BUSINESS/JOB_TITLE || /GOVERNMENT/GOVERNMENT_OFFICE_OR_TITLE || /GOVERNMENT/GOVERNMENT_OFFICE_CATEGORY
/LOCATION/STRUCTURE/AIRPORT	:= /AVIATION/AIRPORT	
/LOCATION/STRUCTURE	:= /ARCHITECTURE/BUILDING	
/LOCATION/STRUCTURE/HOTEL := /TRAVEL/HOTEL	
/LOCATION/STRUCTURE/SPORTS_FACILITY := /SPORTS/SPORTS_FACILITY	
/LOCATION/GEOGRAPHY/BODY_OF_WATER := /GEOGRAPHY/BODY_OF_WATER	
/LOCATION/GEOGRAPHY/MOUNTAIN := /GEOGRAPHY/MOUNTAIN	
/LOCATION/GEOGRAPHY := /GEOGRAPHY/*	
/LOCATION/TRANSIT/BRIDGE := /TRANSPORTATION/BRIDGE	
/LOCATION/TRANSIT/RAILWAY := /RAIL/RAILWAY	
/LOCATION/TRANSIT/ROAD := /TRANSPORTATION/ROAD	
/LOCATION/CITY := /LOCATION/CITYTOWN	
/LOCATION/COUNTRY := /LOCATION/COUNTRY	
/LOCATION/PARK := /AMUSEMENT_PARKS/PARK || /BASE/USNATIONALPARKS/US_NATIONAL_PARK
/LOCATION := /LOCATION/LOCATION	
/ORGANIZATION := /ORGANIZATION/ORGANIZATION_TYPE	|| /ORGANIZATION/ORGANIZATION
/ORGANIZATION/COMPANY/NEWS := /BASE/NEWSEVENTS/NEWS_REPORTING_ORGANISATION || /BOOK/PUBLISHING_COMPANY
/ORGANIZATION/COMPANY/BROADCAST := /BROADCAST/PRODUCER	
/ORGANIZATION/COMPANY := /BUSINESS/EMPLOYER	
/ORGANIZATION/EDUCATION := /EDUCATION/ACADEMIC_INSTITUTION	|| /EDUCATION/EDUCATIONAL_INSTITUTION
/ORGANIZATION/GOVERNMENT := /GOVERNMENT/GOVERNMENT_AGENCY || /GOVERNMENT/GOVERNMENT || /GOVERNMENT/GOVERNMENTAL_BODY
/ORGANIZATION/MILITARY := /MILITARY/MILITARY_UNIT	
/ORGANIZATION/POLITICAL_PARTY := /GOVERNMENT/POLITICAL_PARTY	
/ORGANIZATION/SPORTS_TEAM := /SPORTS/SPORTS_TEAM	
/ORGANIZATION/STOCK_EXCHANGE := /FINANCE/STOCK_EXCHANGE	
/OTHER/ART/BROADCAST := /TV/TV_PROGRAM	
/OTHER/ART/FILM	:= /FILM/FILM	
/OTHER/ART/MUSIC := /MUSIC/ALBUM || /MUSIC/COMPOSITION	
/OTHER/ART/STAGE := /THEATER/PLAY	||  /OPERA/OPERA
/OTHER/ART/WRITING := /BOOK/WRITTEN_WORK || /BOOK/SHORT_STORY || /BOOK/POEM || /BOOK/LITERARY_SERIES || /BOOK/PUBLICATION
/OTHER/EVENT := /TIME/EVENT	
/OTHER/EVENT/HOLIDAY := /TIME/HOLIDAY	
/OTHER/EVENT/VIOLENT_CONFLICT := /MILITARY/MILITARY_CONFLICT	
/OTHER/HEALTH/TREATMENT := /MEDICINE/MEDICAL_TREATMENT	
/OTHER/AWARD := /AWARD/AWARD	
/OTHER/BODY_PART := /MEDICINE/ANATOMICAL_STRUCTURE	
/OTHER/CURRENCY	:= /FINANCE/CURRENCY	
/OTHER/LIVING_THING/ANIMAL := /BIOLOGY/ANIMAL	
/OTHER/LIVING_THING	:= /BASE/PLANTS/PLANT	
/OTHER/PRODUCT/WEAPON := /LAW/INVENTION	
/OTHER/PRODUCT/VEHICLE := /AUTOMOTIVE/MODEL || /AVIATION/AIRCRAFT_MODEL
/OTHER/PRODUCT/COMPUTER := /COMPUTER/*	
/OTHER/PRODUCT/SOFTWARE := /COMPUTER/SOFTWARE	
/OTHER/FOOD	:= /FOOD/FOOD	
/OTHER/RELIGION	:= /RELIGION/RELIGION	
/OTHER/HERITAGE	:= /PEOPLE/ETHNICITY	
/OTHER/LEGAL := /USER/SPROCKETONLINE/ECONOMICS/LEGISLATION	|| /USER/TSEGARAN/LEGAL/ACT_OF_CONGRESS || /USER/SKUD/LEGAL/TREATY || /LAW/CONSTITUTIONAL_AMENDMENT 
/OTHER := !ALL_TYPES_EXLUCDING_OTHER* || /OTHER*
\end{lstlisting}

\begin{lstlisting}[caption={Type definition used of the \ontonotes\ set.}]
/PERSON := /PEOPLE/* $\with \with$ /MUSIC/ARTIST
/LOC := /LOCATION/LOCATION
/ORG := /ORGANIZATION/* || /GOVERNMENT/GOVERNMENT_BODY || /BUSINESS/EMPLOYER || /BOOK/NEWSPAPER || /RELIGION/RELIGION || /MILITARY/MILITARY_COMBATANT
\end{lstlisting}

\begin{lstlisting}[caption={Type definition used of the \muc\ set.}]
/PER := /PEOPLE/* $\with \with$ /MUSIC/ARTIST
/LOC := /LOCATION/LOCATION
/ORG := /ORGANIZATION/* || /GOVERNMENT/GOVERNMENT_BODY || /BOOK/NEWSPAPER || /RELIGION/RELIGION
\end{lstlisting}

\begin{lstlisting}[caption={Type definition used of the \conll\ set.}]
/PER := /PEOPLE/PERSON
/LOC := /LOCATION/LOCATION
/ORG := /ORGANIZATION/ORGANIZATION
\end{lstlisting}

\begin{lstlisting}[caption={Type definition used of the \bionlp\ set.}]
/BACTERIA := /*/MICROORGANISM/*
\end{lstlisting}

\end{document}